\documentclass[11pt,a4paper]{article}
\usepackage[hyperref]{acl2020}
\usepackage{times}
\usepackage{latexsym}

\usepackage{microtype}

\usepackage{amsmath,amsfonts,bm}

\def\eqref#1{equation~\ref{#1}}

\def\1{\bm{1}}

\def\rvc{{\mathbf{c}}}

\def\rvh{{\mathbf{h}}}

\def\rvs{{\mathbf{s}}}
\def\rvt{{\mathbf{t}}}

\def\rvx{{\mathbf{x}}}
\def\rvy{{\mathbf{y}}}

\def\ervc{{\textnormal{c}}}

\def\ervk{{\textnormal{k}}}

\def\ervq{{\textnormal{q}}}

\def\ervs{{\textnormal{s}}}
\def\ervt{{\textnormal{t}}}

\def\ervv{{\textnormal{v}}}

\def\ervx{{\textnormal{x}}}
\def\ervy{{\textnormal{y}}}

\def\vx{{\bm{x}}}

\def\vz{{\bm{z}}}

\def\mX{{\bm{X}}}

\DeclareMathAlphabet{\mathsfit}{\encodingdefault}{\sfdefault}{m}{sl}
\SetMathAlphabet{\mathsfit}{bold}{\encodingdefault}{\sfdefault}{bx}{n}

\newcommand{\sigmoid}{\sigma}

\DeclareMathOperator*{\argmax}{arg\,max}

\usepackage[textsize=tiny]{todonotes}
\usepackage{amsfonts}
\usepackage{amsmath}
\usepackage{amssymb}
\usepackage{bbm}
\usepackage[hang,flushmargin]{footmisc}
\usepackage{lipsum}
\usepackage{makecell}
\usepackage{multirow}
\usepackage{subfigure}
\usepackage{threeparttable}
\usepackage[normalem]{ulem}

\newcommand{\ts}{\textsuperscript}

\makeatletter
\newcommand\footnoteref[1]{\protected@xdef\@thefnmark{\ref{#1}}\@footnotemark}
\makeatother

\definecolor{red1}{RGB}{225, 123, 123}
\definecolor{green1}{HTML}{75AE62}

\aclfinalcopy 
\def\aclpaperid{760} 

\title{Regularized Context Gates on Transformer for Machine Translation}

\author{%
    Xintong Li\textsuperscript{1},
    Lemao Liu\textsuperscript{2},
    Rui Wang\textsuperscript{3},
    Guoping Huang\textsuperscript{2},
    Max Meng\textsuperscript{1} \\\\
	\textsuperscript{1}The Chinese University of Hong Kong $\quad$
	\textsuperscript{2}Tencent AI Lab \\
    \textsuperscript{3}National Institute of Information and Communications Technology \\
    znculee@gmail.com $\quad$
    redmondliu@tencent.com $\quad$
    wangrui@nict.go.jp \\
    donkeyhuang@tencent.com $\quad$
    max.meng@cuhk.edu.hk
}

\date{}

\begin{document}
\maketitle
\begin{abstract}

%
Context gates are effective to control the contributions from the source and
target contexts in the recurrent neural network (RNN) based neural machine
translation (NMT).
However, it is challenging to extend them into the advanced Transformer
architecture, which is more complicated than RNN.
This paper first provides a method to identify source and target contexts and
then introduce a gate mechanism to control the source and target contributions
in Transformer.
In addition, to further reduce the bias problem in the gate mechanism, this
paper proposes a regularization method to guide the learning of the gates with
supervision automatically generated using pointwise mutual information.
Extensive experiments on 4 translation datasets demonstrate that the proposed
model obtains an averaged gain of 1.0 BLEU score over a strong Transformer
baseline.

\end{abstract}


\section{Introduction}

An essence to modeling translation is how to learn an effective context from a
sentence pair. Statistical machine translation (SMT) models the source context
from the source-side of a translation model and models the target context from
a target-side language model~\cite{koehn2003statistical, koehn2009statistical,
chiang2005hierarchical}. These two models are trained independently. On the
contrary, neural machine translation (NMT) advocates a unified manner to
jointly learn source and target context using an encoder-decoder framework with
an attention mechanism, leading to substantial gains over SMT in translation
quality~\cite{sutskever2014sequence, bahdanau2014neural,
gehring2017convolutional, vaswani2017attention}. Prior work on attention
mechanism~\citep{luong2015effective, liu2016neural, mi2016supervised,
chen2018syntax, li2018target, elbayad2018pervasive, yang2020neural} have shown
a better context representation is helpful to translation performance.

\begin{figure}[t]
    \centering
    \includegraphics[width=\columnwidth]{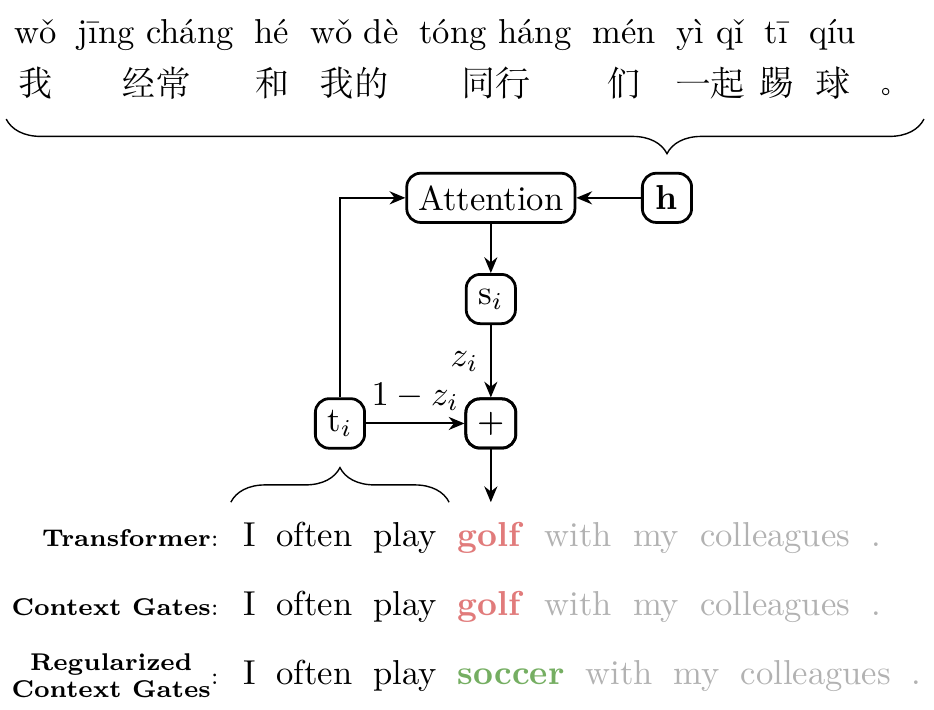}
    \caption{
        A running example to raise the context control problem. Both original
        and context gated Transformer obtain an unfaithful translation by
        wrongly translate ``\textit{t\= i q\' iu}'' into ``\textit{play
        {\color{red1} golf}}'' because referring too much target context. By
        regularizing the context gates, the purposed method corrects the
        translation of ``\textit{t\= i q\' iu}'' into ``\textit{play
        {\color{green1} soccer}}''. The light font denotes the target words to
        be translated in the future. For original Transformer, the source and
        target context are added directly without any rebalancing.
    }
    \label{fig:example}
\end{figure}

However, a standard NMT system is incapable of effectively controlling the
contributions from source and target contexts~\citep{he2018layer} to deliver
highly adequate translations as shown in Figure~\ref{fig:example}. As a result,
\citet{tu2017context} carefully designed context gates to dynamically control
the influence from source and target contexts and observed significant
improvements in the recurrent neural network (RNN) based NMT.  Although
Transformer~\cite{vaswani2017attention} delivers significant gains over RNN for
translation, there are still one third translation errors related to context
control problem as described in Section~\ref{subsec:error}. Obviously, it is
feasible to extend the context gates in RNN based NMT into Transformer, but an
obstacle to accomplishing this goal is the complicated architecture in
Transformer, where the source and target words are tightly coupled. Thus, it is
challenging to put context gates into practice in Transformer.

In this paper, under the Transformer architecture, we firstly provide a way to
define the source and target contexts and then obtain our model by combining
both source and target contexts with context gates, which actually induces a
probabilistic model indicating whether the next generated word is contributed
from the source or target sentence~\citep{li2019word}. In our preliminary
experiments, this model only achieves modest gains over Transformer because the
context selection error reduction is very limited as described in
Section~\ref{subsec:error}.  To further address this issue, we propose a
probabilistic model whose loss function is derived from external supervision as
regularization for the context gates. This probabilistic model is jointly
trained with the context gates in NMT. As it is too costly to annotate this
supervision for a large-scale training corpus manually, we instead propose a
simple yet effective method to automatically generate supervision using
pointwise mutual information, inspired by word collocation
~\cite{bouma2009normalized}.  In this way, the resulting NMT model is capable
of controlling the contributions from source and target contexts effectively.

We conduct extensive experiments on 4 benchmark datasets, and experimental
results demonstrate that the proposed gated model obtains an averaged
improvement of 1.0 BLEU point over corresponding strong Transformer baselines.
In addition, we design a novel analysis to show that the improvement of
translation performance is indeed caused by relieving the problem of wrongly
focusing on the source or target context.


\section{Methodology}

Given a source sentence $\rvx = \langle \ervx_1, \cdots, \ervx_{|\rvx|}\rangle$
and a target sentence $\rvy = \langle \ervy_1, \cdots, \ervy_{|\rvy|}\rangle$,
our proposed model is defined by the following conditional probability under
the Transformer architecture:~\footnote{Throughout this paper, a variable in
bold font such as $\rvx$ denotes a sequence while regular font such as $\ervx$
denotes an element which may be a scalar $x$, vector $\vx$ or matrix $\mX$.}
\begin{equation}\begin{array}{rcl}
    P \left(\rvy \mid \rvx\right) & = & \prod \limits _{i=1}^{|\rvy|}
    P \left(\ervy_i \mid \rvy_{<i}, \rvx\right) =
    \prod \limits _{i=1}^{|\rvy|} P \left(\ervy_i \mid \ervc_i^L\right),
    \label{eq:nmt}
\end{array}\end{equation}
where $\rvy_{<i} = \langle \ervy_1, \dots, \ervy_{i-1}\rangle$ denotes a prefix
of $\rvy$ with length $i-1$, and $\ervc_i^L$ denotes the $L$\ts{th} layer
context in the decoder with $L$ layers which is obtained from the
representation of $\rvy_{<i}$ and $\rvh^L$, i.e., the top layer hidden
representation of $\rvx$, similar to the original Transformer. To finish the
overall definition of our model in \eqref{eq:nmt}, we will expand the
definition $\ervc_i^L$ based on context gates in the following subsections.

\subsection{Context Gated Transformer}
\label{subsec:cg_transformer}

To develop context gates for our model, it is necessary to define the source
and target contexts at first. Unlike the case in RNN, the source sentence
$\rvx$ and the target prefix $\rvy_{<i}$ are tightly coupled in our model, and
thus it is not trivial to define the source and target contexts.

Suppose the source and target contexts at each layer $l$ are denoted by
$\ervs_i^l$ and $\ervt_i^l$. We recursively define them from $\rvc_{<i}^{l-1}$
as follows.~\footnote{For the base case, $\rvc_{<i}^0$ is word embedding of
$\rvy_{<i}$.}
\begin{equation}\begin{array}{rcl}
    \ervt_i^l & = & \mathrm{rn}\circ\mathrm{ln}\circ\mathrm{att}
                    \left(\ervc_i^{l-1}, \rvc_{<i}^{l-1}\right), \\
    \ervs_i^l & = &\mathrm{ln}\circ\mathrm{att}\left(\ervt_i^l,\rvh^L\right),
\end{array}\end{equation}
where $\circ$ is functional composition, $\mathrm{att} \left(\ervq, \ervk
\ervv\right)$ denotes multiple head attention with $\ervq$ as query, $\ervk$ as
key, $\ervv$ as value, and $\mathrm{rn}$ as a residual
network~\cite{he2016deep}, $\mathrm{ln}$ is layer
normalization~\cite{ba2016layer}, and all parameters are removed for
simplicity.

In order to control the contributions from source or target side, we define
$\ervc_i^l$ by introducing a context gate $\vz_i^l$ to combine $\ervs_i^l$ and
$\ervt_i^l$ as following:
\begin{equation}\begin{array}{rcl}
    \rvc_i^l & = & \mathrm{rn}\circ\mathrm{ln}\circ\mathrm{ff}
    \left((\1-\vz_i^l)\otimes\ervt_i^l+
    \vz_i^l\otimes\ervs_i^l\right)
    \label{eq:ctx}
\end{array}\end{equation}
with
\begin{equation}\begin{array}{rcl}
    \vz_i^l & = & \sigmoid \left( \mathrm{ff} \left( \ervt_i^l \Vert
    \ervs_i^l\right) \right),
    \label{eq:gates}
\end{array}\end{equation}
where \noindent $\textrm{ff}$ denotes a feedforward neural network, $\Vert$
denotes concatenation, $\sigma(\cdot)$ denotes a sigmoid function, and
$\otimes$ denotes an element-wise multiplication. $\vz_i^l$ is a vector
(\citet{tu2017context} reported that a gating vector is better than a gating
scalar). Note that each component in $\vz_i^l$ actually induces a probabilistic
model indicating whether the next generated word $\ervy_i$ is mainly
contributed from the source ($\rvx$) or target sentence ($\rvy_{<i}$) , as
shown in Figure~\ref{fig:example}.

\paragraph{Remark}
It is worth mentioning that our proposed model is similar to the standard
Transformer with boiling down to replacing a residual connection with a high
way connection~\citep{srivastava2015highway, zhang2018improving}: if we replace
$(\1-\vz_i^l) \otimes \ervt_i^l + \vz_i^l \otimes \ervs_i^l$ in \eqref{eq:ctx}
by $\rvt_i^l + \rvs_i^l$, the proposed model is reduced to Transformer.

\subsection{Regularization of Context Gates}

In our preliminary experiments, we found learning context gates from scratch
cannot effectively reduce the context selection errors as described in
Section~\ref{subsec:error}.

To address this issue, we propose a regularization method to guide the learning
of context gates by external supervision $z_i^*$ which is a binary number
representing whether $\ervy_i$ is contributed from either source ($z_i^*=1$) or
target sentence ($z_i^*=0$). Formally, the training objective is defined as
follows:
\begin{multline}
    \ell = - \log P(\rvy\mid \rvx) + \lambda \sum \limits _{l,i} \bigg(
        z_i^*\max ( \bm{0.5} - \vz_i^l, \bm{0} ) \\ +
        (1-z_i^*)\max ( \vz_i^l - \bm{0.5}, \bm{0} ) \bigg),
    \label{eq:reg}
\end{multline}
where $\vz_i^l$ is a context gate defined in \eqref{eq:gates} and $\lambda$
is a hyperparameter to be tuned in experiments. Note that we only regularize
the gates during the training, but we skip the regularization during inference.

Because golden $z_i^*$ are inaccessible for each word $\ervy_i$ in the training
corpus, we ideally have to annotate it manually. However, it is costly for
human to label such a large scale dataset. Instead, we propose an automatic
method to generate its value in practice in the next subsection.

\subsection{Generating Supervision $z_i^*$}

To decide whether $\ervy_i$ is contributed from the source ($\rvx$) or target
sentence ($\rvy_{<i}$)~\citep{li2019word}, a metric to measure the correlation
between a pair of words ($\langle \ervy_i, \ervx_j\rangle$ or $\langle \ervy_i,
\ervy_k\rangle$ for $k<i$) is first required. This is closely related to a
well-studied problem, i.e., word collocation~\cite{liu2009collocation}, and we
simply employ the pointwise mutual information (PMI) to measure the correlation
between a word pair $\langle \mu, \nu\rangle$
following~\citet{,bouma2009normalized}:
\begin{equation}\begin{array}{rcl}
    \mathrm{pmi}\left(\mu,\nu\right) & = &
    \log \frac{P\left(\mu,\nu\right)}{P\left(\mu\right)P\left(\nu\right)} \\
    & = & \log Z +
    \log \frac{C\left(\mu,\nu\right)}{C\left(\mu\right)C\left(\nu\right)},
\end{array}\end{equation}
where $C\left(\mu\right)$ and $C\left(\nu\right)$ are word counts,
$C\left(\mu,\nu\right)$ is the co-occurrence count of words $\mu$ and $\nu$,
and $Z$ is the normalizer, i.e., the total number of all possible
$\left(\mu,\nu\right)$ pairs. To obtain the context gates, we define two types
of PMI according to different $C\left(\mu,\nu\right)$ including two scenarios
as follows.

\paragraph{PMI in the Bilingual Scenario}
For each parallel sentence pair $\langle\rvx,\rvy\rangle$ in training set,
$C\left(\ervy_i,\ervx_j\right)$ is added by one if both $\ervy_i\in\rvy$ and
$\ervx_j\in\rvx$.

\paragraph{PMI in the Monolingual Scenario}
In the translation scenario, only the words in the preceding context of a
target word should be considered. So for any target sentence $\rvy$ in the
training set, $C\left(\ervy_i,\ervy_k\right)$ is added by one if both
$\ervy_i\in\rvy$ and $\ervy_k\in\rvy_{<i}$.

Given the two kinds of PMI for a bilingual sentence $\langle\rvx,\rvy\rangle$,
each $z_i^*$ for each $\ervy_i$ is defined as follows,
\begin{equation}
    z_i^* =
    \mathbbm{1}_{\max_j \mathrm{pmi}\left(\ervy_i, \ervx_j\right)>
    \max_{k<i} \mathrm{pmi}\left(\ervy_i, \ervy_k\right)},
    \label{eq:z*}
\end{equation}
where $\mathbbm{1}_b$ is a binary function valued by 1 if $b$ is true and 0
otherwise. In \eqref{eq:z*}, we employ $\max$ strategy to measure the
correlation between $\ervy_i$ and a sentence ($\rvx$ or $\rvy_{<i}$). Indeed,
it is similar to use the average strategy, but we did not find its gains over
$\max$ in our experiments.


\begin{table*}[htb]\small
    \centering
    \begin{tabular}{c|c|c|c|c|c||c|c|c}
        \multicolumn{2}{c|}{\multirow{2}{*}{\textbf{Models}}} & \multirow{2}{*}{\makecell{\textbf{params}\\$\times10^6$}} & \multicolumn{3}{c||}{ZH$\Rightarrow$EN} & \multirow{2}{*}{EN$\Rightarrow$DE} & \multirow{2}{*}{DE$\Rightarrow$EN} & \multirow{2}{*}{FR$\Rightarrow$EN} \\
        \cline{4-6}
        \multicolumn{1}{c}{}                              &    &      & MT05 & MT06 & MT08 &      &       &  \\
        \hline
        \multicolumn{2}{c|}{RNN based NMT}                & 84 & 30.6 & 31.1 & 23.2 & --   & --   & --   \\
        \multicolumn{2}{c|}{\citet{tu2017context}}        & 88 & 34.1 & 34.8 & 26.2 & --   & --   & --   \\
        \multicolumn{2}{c|}{\citet{vaswani2017attention}} & 65 & --   & --   & --   & 27.3 & --   & --   \\
        \multicolumn{2}{c|}{\citet{ma2018bag}}            & -- & 36.8 & 35.9 & 27.6 & --   & --   & --   \\
        \multicolumn{2}{c|}{\citet{zhao2018addressing}}   & -- & 43.9 & 44.0 & 33.3 & --   & --   & --   \\
        \multicolumn{2}{c|}{\citet{cheng2018towards}}     & -- & 44.0 & 44.4 & 34.9 & --   & --   & --   \\
        \hline
        \hline
        \multicolumn{2}{c|}{Transformer}                  & 74 & 46.9 & 47.4 & 38.3 & 27.4 & 32.2 & 36.8 \\
        \hline
        \multirow{2}{*}{This Work}
            & Context Gates                               & 92 & 47.1 & 47.6 & 39.1 & 27.9 & 32.5 & 37.7 \\
            & Regularized Context Gates                   & 92 & \textbf{47.7} & \textbf{48.3} & \textbf{39.7} & \textbf{28.1} & \textbf{33.0} & \textbf{38.3} \\
    \end{tabular}
    \caption{
        Translation performances (BLEU). The RNN based NMT
        \citep{bahdanau2014neural} is reported from the baseline model in
        \citet{tu2017context}. ``params" shows the number of parameters of
        models when training ZH$\Rightarrow$EN except
        \citet{vaswani2017attention} is for EN$\Rightarrow$DE tasks.
         }
    \label{tab:tp}
\end{table*}

\section{Experiments}

The proposed methods are evaluated on
NIST ZH$\Rightarrow$EN \footnote{LDC2000T50,~LDC2002L27,~LDC2002T01,~LDC2002E18,
                                 LDC2003E07,~LDC2003E14,~LDC2003T17,~LDC2004T07},
WMT14 EN$\Rightarrow$DE \footnote{WMT14: http://www.statmt.org/wmt14/},
IWSLT14 DE$\Rightarrow$EN \footnote{IWSLT14: http://workshop2014.iwslt.org/} and
IWSLT17 FR$\Rightarrow$EN \footnote{IWSLT17: http://workshop2017.iwslt.org/} tasks.
To make our NMT models capable of open-vocabulary translation, all datasets are
preprocessed with Byte Pair Encoding \citep{sennrich2015neural}. All proposed
methods are implemented on top of Transformer \cite{vaswani2017attention} which
is the state-of-the-art NMT system. Case-insensitive BLEU score
\citep{papineni2002bleu} is used to evaluate translation quality of
ZH$\Rightarrow$EN, DE$\Rightarrow$EN and FR$\Rightarrow$EN. For the fair
comparison with the related work, EN$\Rightarrow$DE is evaluated with
case-sensitive BLEU score. Setup details are described in
Appendix~\ref{sec:setup}.

\begin{table}[htb]\small
    \centering
    \begin{threeparttable}
    \begin{tabular}{c|c|c|c|c|c}
        $\lambda$     & 0.1  & 0.5  & 1             & 2    & 10 \\
        \hline
        \textbf{BLEU} & 32.7 & 32.6 & \textbf{33.0} & 32.7 & 32.6
    \end{tabular}
    \begin{tablenotes}
        \scriptsize
        \item[*] Results are measured on DE$\Rightarrow$EN task.
    \end{tablenotes}
    \end{threeparttable}
    \caption{Translation performance over different regularization
             coefficient $\lambda$.}
    \label{tab:lambda}
\end{table}

\subsection{Tuning Regularization Coefficient}

In the beginning of our experiments, we tune the regularization coefficient
$\lambda$ on the DE$\Rightarrow$EN task. Table~\ref{tab:lambda} shows the
robustness of $\lambda$, because the translation performance only fluctuates
slightly over various $\lambda$. In particular, the best performance is
achieved when $\lambda=1$, which is the default setting throughout this paper.

\subsection{Translation Performance}
\label{subsec:tp}

Table~\ref{tab:tp} shows the translation quality of our methods in BLEU. Our
observations are as follows:

1) The performance of our implementation of the Transformer is slightly higher
than \citet{vaswani2017attention}, which indicates we are in a fair comparison.

2) The proposed Context Gates achieves modest improvement over the baseline. As
we mentioned in Section~\ref{subsec:cg_transformer}, the structure of RNN based
NMT is quite different from the Transformer. Therefore, naively introducing the
gate mechanism to the Transformer without adaptation does not obtain similar
gains as it does in RNN based NMT.

3) The proposed Regularized Context Gates improves nearly 1.0 BLEU score over
the baseline and outperforms all existing related work. This indicates that
the regularization can make context gates more effective in relieving the
context control problem as discussed following.

\subsection{Error Analysis}
\label{subsec:error}

To explain the success of Regularized Context Gates, we analyze the error rates
of translation and  context selection. Given a sentence pair $\rvx$ and $\rvy$,
the forced decoding translation error is defined as $P \left(\ervy_i \mid
\rvy_{<i}, \rvx\right) < P \left(\hat{\ervy}_i \mid \rvy_{<i}, \rvx\right)$,
where $\hat{\ervy}_i \triangleq \argmax _\ervv P \left(\ervv \mid \rvy_{<i},
\rvx\right)$ and $\ervv$ denotes any token in the vocabulary. The context
selection error is defined as $z_i^*(\ervy_i) \neq z_i^*(\hat{\ervy}_i)$, where
$z_i^*$ is defined in \eqref{eq:z*}. Note that a context selection error must
be a translation error but the opposite is not true. The example shown in
Figure~\ref{fig:example} also demonstrates a context selection error indicating
the translation error is related with the bad context selection.

\begin{table}[htb]\small
    \newcommand{\cete}{$\text{\textbf{CE}}/\text{\textbf{FE}}$}
    \centering
    \begin{threeparttable}
    \begin{tabular}{c|c|c|c}
        \textbf{Models}           & \textbf{FER}  & \textbf{CER}  & \cete         \\
        \hline
        Transformer               & 40.5          & 13.8          & 33.9          \\
        Context Gates             & 40.5          & 13.7          & 33.7          \\
        Regularized Context Gates & \textbf{40.0} & \textbf{13.4} & \textbf{33.4} \\
    \end{tabular}
    \begin{tablenotes}
        \scriptsize
        \item[*] Results are measured on MT08 of ZH$\Rightarrow$EN task.
    \end{tablenotes}
    \end{threeparttable}
    \caption{
        Forced decoding translation error rate (\textbf{FER}), context
        selection error rate (\textbf{CER}) and the proportion of context
        selection errors over forced decoding translation errors (\cete) of the
        original and context gated Transformer with or without regularization.
    }
    \label{tab:err}
    \let\cete\undefined
\end{table}

As shown in Table~\ref{tab:err}, the Regularized Context Gates significantly
reduce the translation error by avoiding the context selection error. The
Context Gates are also able to avoid few context selection error but cannot
make a notable improvement in translation performance. It is worth to note that
there is approximately one third translation error is related to context
selection error. The Regularized Context Gates indeed alleviate this severe
problem by effectively rebalancing of source and target context for
translation.

\subsection{Statistics of Context Gates}
\label{subsec:stat_cg}

\begin{table}[htb]\small
    \centering
    \begin{threeparttable}
    \begin{tabular}{c|c|c}
        \textbf{Models}           & \textbf{Mean} & \textbf{Variance} \\
        \hline
        Context Gates             & 0.38          & 0.10         \\
        Regularized Context Gates & 0.51          & 0.13         \\
    \end{tabular}
    \begin{tablenotes}
        \scriptsize
        \item[*] Results are measured on MT08 of ZH$\Rightarrow$EN task.
    \end{tablenotes}
    \end{threeparttable}
    \caption{Mean and variance of context gates}
    \label{tab:mv}
\end{table}

Table~\ref{tab:mv} summarizes the mean and variance of each context gate (every
dimension of the context gate vectors) over the MT08 test set. It shows that
learning context gates freely from scratch tends to pay more attention to
target context (0.38 $<$ 0.5), which means the model tends to trust its
language model more than the source context, and we call this context imbalance
bias of the freely learned context gate. Specifically, this bias will make the
translation unfaithful for some source tokens. As shown in Table~\ref{tab:mv},
the Regularized Context Gates demonstrates more balanced behavior
(0.51$\approx$0.5) over the source and target context with similar variance.

\subsection{Regularization in Different Layers}

To investigate the sensitivity of choosing different layers for regularization,
we only regularize the context gate in every single layer.
Table~\ref{tab:diffly} shows that there is no significant performance
difference, but all single layer regularized context gate models are slightly
inferior to the model, which regularizes all the gates. Moreover, since nearly
no computation overhead is introduced and for design simplicity, we adopt
regularizing all the layers.

\begin{table}[htb]\small
    \centering
    \begin{threeparttable}
    \begin{tabular}{c|c|c|c|c|c|c}
        Layers        & N/A  & 1    & 2    & 3    & 4    & ALL \\
        \hline
        \textbf{BLEU} & 32.5 & 32.8 & 32.7 & 32.5 & 32.3 & 33.0
    \end{tabular}
    \begin{tablenotes}
        \scriptsize
        \item[*] Results are measured on DE$\Rightarrow$EN task.
    \end{tablenotes}
    \end{threeparttable}
    \caption{
        Regularize context gates on different layers.``N/A" indicates
        regularization is not added. ``ALL" indicates regularization is added
        to all the layers.
    }
    \label{tab:diffly}
\end{table}

\subsection{Effects on Long Sentences}

\begin{figure}[htb]
    \centering
    \includegraphics[width=\columnwidth]{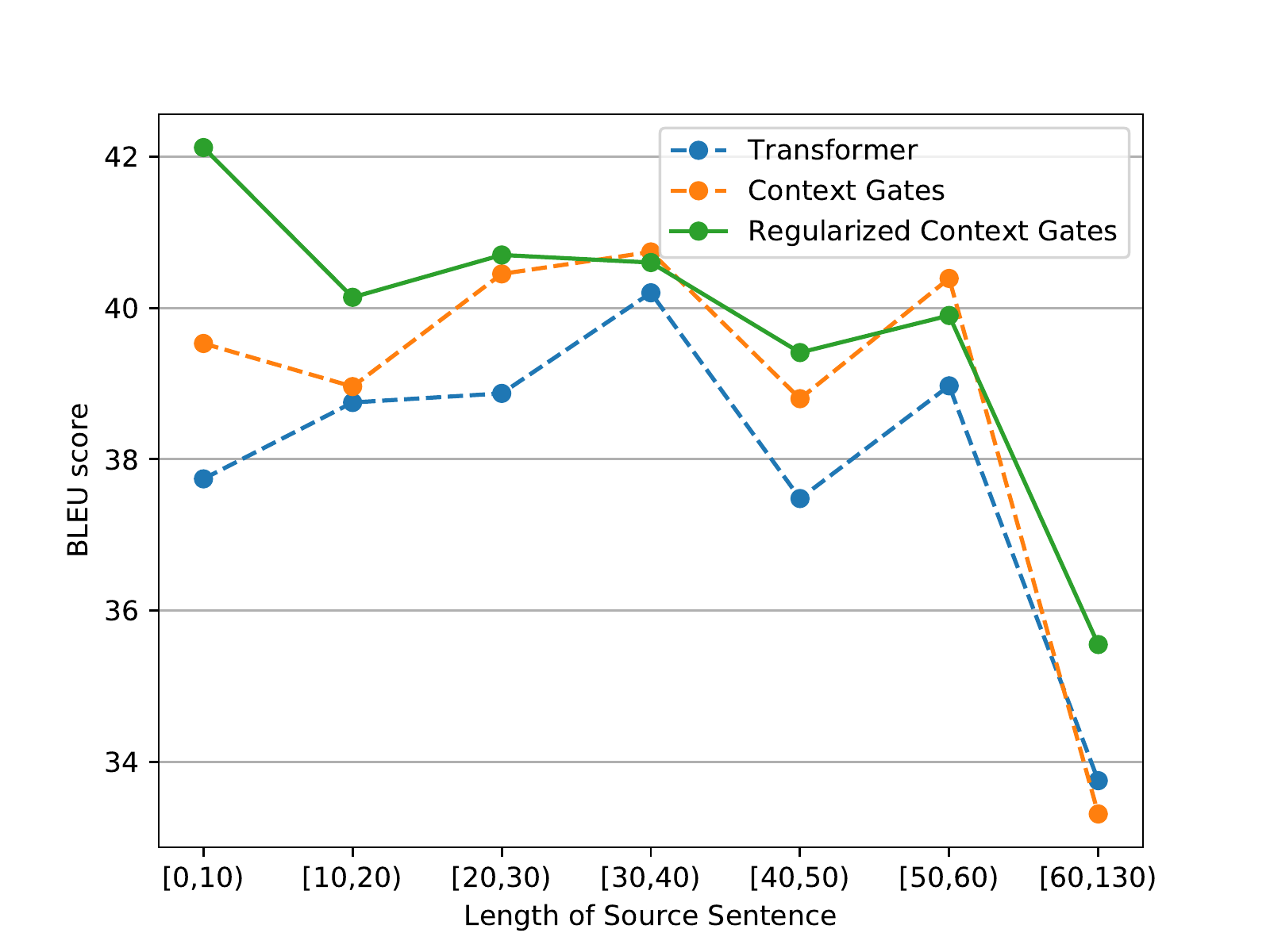}
    \caption{Translation performance on MT08 test set with respect to
             different lengths of source sentence. Regularized Context Gates
             significantly improves the translation of short and long
             sentences.}
    \label{fig:difflen}
\end{figure}

In \citet{tu2017context}, context gates alleviate the problem of long sentence
translation of attentional RNN based system \cite{bahdanau2014neural}. We
follow \citet{tu2017context} and compare the translation performances according
to different lengths of the sentences. As shown in Figure~\ref{fig:difflen}, we
find Context Gates does not improve the translation of long sentences but
translate short sentences better. Fortunately, the Regularized Context Gates
indeed significantly improves the translation for both short sentences and long
sentences.

\section{Conclusions}

This paper transplants context gates from the RNN based NMT to the Transformer
to control the source and target context for translation. We find that context
gates only modestly improve the translation quality of the Transformer, because
learning context gates freely from scratch is more challenging for the
Transformer with the complicated structure than for RNN. Based on this
observation, we propose a regularization method to guide the learning of
context gates with an effective way to generate supervision from training data.
Experimental results show the regularized context gates can significantly
improve translation performances over different translation tasks even though
the context control problem is only slightly relieved. In the future, we
believe more work on alleviating context control problem has the potential to
improve translation performance as quantified in Table~\ref{tab:err}.

\bibliography{ref.bib}
\bibliographystyle{acl_natbib}

\newpage \clearpage \appendix

\section{Details of Data and Implementation}
\label{sec:setup}

The training data for ZH$\Rightarrow$EN task consists of 1.8M sentence pairs.
The development set is chosen as NIST02 and test sets are NIST05, 06, 08. For
EN$\Rightarrow$DE task, its training data contains 4.6M sentences pairs. Both
FR$\Rightarrow$EN and DE$\Rightarrow$EN tasks contain around 0.2M sentence
pairs. For ZH$\Rightarrow$EN and EN$\Rightarrow$DE tasks, the joint vocabulary
is built with 32K BPE merge operations, and for DE$\Rightarrow$EN and
FR$\Rightarrow$EN tasks it is built with 16K merge operations.

Our implementation of context gates and the regularization are based on
Transformer, implemented by THUMT \citep{zhang2017thumt}.  For ZH$\Rightarrow$EN and
EN$\Rightarrow$DE tasks, only the sentences of length up to 256 tokens are used
with no more than $2^{15}$ tokens in a batch. The dimension of both word
embeddings and hidden size are 512. Both encoder and decoder have 6 layers and
adopt multi-head attention with 8 heads.  For FR$\Rightarrow$EN and
DE$\Rightarrow$EN tasks, we use a smaller model with 4 layers and 4 heads, and
both the embedding size and the hidden size is 256. The training batch contains
no more than $2^{12}$ tokens.  For all tasks, the beam size for decoding is 4,
and the loss function is optimized with Adam, where $\beta_1=0.9$, $\beta_2=0.98$
and $\epsilon=10^{-9}$.




\end{document}